\documentclass{elsarticle}
\usepackage{algorithm}
\usepackage{algorithmic}

\usepackage{amsmath}
\usepackage{amssymb}
\usepackage{amsthm}
    \theoremstyle{plain}
    \newtheorem{definition}{Definition}[section]
\usepackage{graphicx}
\usepackage{hyperref}
\usepackage[caption=false]{subfig}
\begin{document}
\begin{frontmatter}
\title{A Fixed point view: A Model-Based Clustering Framework}

\author[HUST]{Jianhao Ding\fnref{DJH}}
\author[HUST]{Lansheng Han\corref{HLS}}

\begin{abstract}
With the inflation of the data, clustering analysis, as a branch of unsupervised learning, lacks unified understanding and application of its mathematical law. Based on the view of fixed point, this paper restates the model-based clustering and proposes a unified clustering framework. In order to find fixed points as cluster centers, the framework iteratively constructs the contraction map, which strongly reveals the convergence mechanism and interconnections among algorithms. By specifying a contraction map, Gaussian mixture model (GMM) can be mapped to the framework as an application. We hope the fixed point framework will help the design of future clustering algorithms.
\end{abstract}


\address[HUST]{School of Computer Science and Technology, Huazhong University of Science and Technology, Wuhan 430074, China}
\fntext[DJH]{Jianhao Ding, Undergraduate of School of Computer Science and Technology, Huazhong University of Science and Technology.
Email: dingjh1998@hust.edu.cn.}
\cortext[HLS]{Lansheng Han, Professor of Information Security, School of Computer Science and Technology, Huazhong University of Science and Technology.
Email: 1998010309@hust.edu.cn. Academic Areas: Information Security, Security of Network, Security of Big Data.}

\end{frontmatter}

\section{Introduction}\label{sec:intro}
Nowadays, we are in the Big Data Era, human society can produce tens of thousands of unstructured or semi-structured data in a second. However, not all of the data are representative and meaningful, so the analysis and disposal of large-scale data occupies an increasingly important position in scientific research and social life \cite{Sowmya2017}. Cluster analysis is an important unsupervised learning method in machine learning. Its basic idea is grouping a set of objects into clusters, in a way that objects in the same cluster share more similarity than those from separated clusters, in terms of distances of a certain space.

In the evolution of clustering, due to the differences of data types and clustering strategies, cluster analysis can be divided into two main branches, namely, traditional clustering algorithms and modern clustering algorithms. Traditional clustering algorithms include clustering algorithm based on partition, density, model, fuzzy theory and so on \cite{Xu2015,NERURKAR2018770}. In contrast, modern clustering algorithms are mainly presented from the perspective of modern scientific research or swarm intelligence.

With respect of representativeness and inclusiveness of model-based clustering \cite{mcnicholas2016model}, this paper focuses on the uniform characteristics of model-based clustering algorithms. Such algorithms require a priori. Thus, the iterative scheme varies with the priori. Probabilistic clustering model, for example, Expectation Maximization Algorithm (EM) and Gaussian mixture model algorithm (GMM), use maximum likelihood estimation to obtain the model parameters \cite{Jin2017,Reynolds2009}; algorithms based on neural network get the clusters by adjusting weights of neurons in neural network. Typical algorithms are self-organizing feature map (SOM), Neural-Gas algorithm and ART \cite{CARPENTER1990129,Carpenter198712,CARPENTER198754}. For the same dataset, different priori models lead to different clustering results. As model-based clustering algorithms need to figure out the model parameters, they usually have a high time complexity. But the advantage is that the convergent model can explain the distribution features of the data.

\subsection{Motivation and Contributions}\label{subsec:moti}
There are a variety of model- based clustering algorithms. The hypothesis of similarity and the methodology of parameter updating are not necessarily interconnected. Sometimes, the iterative method can be used as an independent module. For example, the GMM algorithm and the EM algorithm share the same iterative method to maximize parameters before convergence. The goal of convergence is to obtain explicit model parameters, which also applies to the neural network model. By studying the convergent model and visualizing the membership degree of each point on the data space $R^n$, the visualization illustrates the point that most of the data can be represented by a local model, and the distribution of membership degree exists convex structure. In other words, there may be more than one extremum in the convex hull of any cluster. This inspires us to reexamine the problem from the view of fixed point.

The existing framework to date has tended to focus on data distribution or partition. Extensive research has rarely been carried out on which framework to apply in order to achieve better clustering and how to discriminate the convergence of algorithm \cite{Zhong2003,Teboulle2007}. The convergence of a clustering algorithm corresponds to the finite time to complete the iterative process. Different clustering models need a framework to better unify the workflow of model-based clustering algorithms. Besides, a working model can be regarded as a contraction map in nature, which may reveal common features of the similarity and the dissimilarity of the data. The updating of contraction map compresses the image space in the iteration of clustering, and thus completes the algorithm.

The clustering algorithm is generally more dependent on cluster shape. The choice of model and also fixed the shape of the cluster. Gaussian mixture model, for example, is a clustering model based on Gaussian distribution, so the shape of clusters is typically a hyper ellipsoid. The Gaussian distribution hypothesis is often appropriate for some datasets, but it is invalid for datasets without Gaussian distribution feature. While in hierarchical clustering, the data distribution hypothesis is weakened, so the algorithm can identify more complex cluster shapes. By emphasizing the similarity or relations with neighboring data, there exists a unified framework, which is less sensitive to the cluster shape and satisfies the requirements of various cluster.

Besides, a typical algorithm can only be classified into one category of clustering. Considering the global or local linear properties, the strong generalization of neural networks can support a clustering framework. Nowadays, neural networks perform well in supervised learning using their forward and back propagation features \cite{Scellier2017}. Self-organizing feature map and its derivations are representative of unsupervised learning methods in neural networks \cite{Carpenter198712,CARPENTER198754}. These algorithms discuss and study the lateral propagation of neural networks. But the order of the neurons in physical implementation is not well embodied, which contributes to a lower efficiency.

In this paper, we present a unified clustering framework based on fixed point. The framework constructs the contraction map by compressing space, and determine the number of clusters by the fixed point. The new framework not only unifies the clustering in theory, but also gives another reasonable explanation for the convergence of clustering algorithm. And this may reveals the inherent laws of itself, rather than simply clustering under specific purposes. Thus, the framework provides a theoretical basis for the construction of new clustering methods.

The paper contributes in the following ways:

\begin{enumerate}[1)]
\item We restate the model-based clustering model.
\item We propose a unified framework from the view of fixed point.
\item We map the GMM algorithm to the proposed framework by specifying a contraction map.
\end{enumerate}

\subsection{Organization}\label{subsec:organ}
The rest of the paper is organized as follows. In Section \ref{sec:related}, axiomatic clustering and some relevant unified clustering framework are reviewed. In Section \ref{sec:unified}, a unified clustering framework from the view of fixed point is presented. Section \ref{sec:extended} discussed the relation between the proposed framework and GMM. Finally, Section \ref{sec:conclusion} summarizes the paper and discusses the future work.

\section{Related Works}\label{sec:related}
Many researchers have expressed their perspective on the issue of unified clustering framework. Jon Kleinberg proposed axiomatic clustering in 2003. He summarized the cluster as \textbf{three axioms}, namely \textbf{Scale-Invariance}, \textbf{Richness} and \textbf{Consistency}. Scale-Invariance requires the partitions are invariant to linear transformations \cite{Kleinberg2003}. Richness requires guarantee of surjection between distance $d$ and partition $\Gamma$. Consistency ensures the invariance of clusters after the measure is reconstructed in term of the similarity between the clusters. In 2009, Zadeh et al. considered to replace the prior three axioms with three weaker axioms, namely Scale-Invariance, Order Consistency and k-Richness \cite{Zadeh2009}. The concept of axiomatic clustering has strong theoretical background, but it is very limited in practical application. On the one hand, algorithms which fully satisfy three axioms are difficult to construct. Zadeh \cite{Zadeh2009}, Bandyopadhyay \cite{Bandyopadhyay2016} and other researchers have tried to construct the algorithm satisfying axioms. On the other hand, Klopotek in \cite{KlopotekK2017} analyzed the association between K-means and Kleinberg's axioms, and pointed out that the absence of cluster shape during the construction of axioms will result in the loss of the validity of the axiomatic clustering.

Axiomatic clustering is the cornerstone of unified clustering framework. More researchers have devoted to studying the framework for some clustering methods. Zhong et al. developed a unified framework for model-based clustering. The framework studied the similarity of data from the perspective of bipartite graph, where data space $\chi$ and cluster model $M$ constitute surjection. The clustering model is further separated into partitional clustering and hierarchical clustering due to the data organization \cite{Zhong2003}. However, this framework ignores the similarity between the data, which will result in merely the characterization of relationship between the components. Based on the optimization theory in convex analysis, Teboulle utilized support and asymptotic functions to construct a continuous framework for clustering in \cite{Teboulle2007}. Li and Vidal formalized an optimization framework to unified the subspace clustering methods. In their framework, affinity matrix is learned. Subspace structured norm and structured sparse subspace $L1$ norm is brought up. Based on the new norm, the spectral clustering is combined to improve the clustering accuracy \cite{Li2015}. In view of information deficiency in spectral clustering, Yang and Shen et al. put forward a unified framework. It is the basis of normal mode that distinguish spectral clustering with other clustering algorithms. Yang et al. replaced the $L2$ Frobenius norm loss with the $L_{2,p}$ norm loss for clustering to improve the generalization ability of the model \cite{Yang2016}.

Each cluster framework mentioned above tends to focus on some specified structural features of the data, with the convergence feature weakened. Nowadays, conditions for a qualitative change in big data is approaching, due to the complexity and volume of big data. Correspondingly, with the clarity of data hierarchy and the increasing time complexity of algorithms, a new framework needs to complement the advantages of different types of clustering. In other words, the research of clustering algorithms should not merely focus on materialization. And the internal relations and rules, instead, need to be extracted, which is the main motivation of this article.

\section{A Unified Framework for Model-based Clustering}\label{sec:unified}
In this section, we propose a new unified framework for model-based clustering from the view of fixed point. The existing framework has many theoretical proofs, but there is no consensus on the structure of the cluster. The framework presented in this paper gives a constructive opinion.
\subsection{Model-based Clustering}\label{subsec:model_based}
With the intent of eliciting the framework, we restate the model-based clustering. Let $X=(X_1,X_2,X_3,\cdots,X_N)$ be a set of $N$ random observed vectors, and the vector is of $L$ dimensions. The universal set $U=Conv(X)\in R^L$ is the convex hull of the $N$ observed vectors. In other words, for all observed vectors $x$, $y$ in the universal set $U$, with $\alpha \in [0,1]$, $\alpha x+(1-\alpha)y \in U$. In general, the model-based clustering has finite mixture models. So the density of $x$ in the model $f(x\vert \pi,\theta)$ is given as:
\begin{equation}\label{eq:model_based}
  f(x\vert \pi,\theta)=\sum_{g=1}^G \pi_g \phi_g (x \vert \theta_g).
\end{equation}

In Equation \ref{eq:model_based}, $G$ is the number of mixed models, and $\pi_g$ is the mixture coefficient, which satisfies $\pi_g>0$ and $\sum_{g=1}^G \pi_g =1$. $\phi_g (x \vert \theta_g)$ represents the density of $x$ in the $g^{th}$ model component. $\theta_g$ is the parameter of the $g^{th}$ component. $\theta=[\theta_1,\theta_2,\cdots,\theta_G]$, $\pi=[\pi_1,\pi_2,\cdots,\pi_G]$. More detailed research on model-based clustering is given in \cite{McNicholas2016}.

The data observation values are correlated with the clusters by the probability. The maximum value of the probabilities can imply the interpretation of the whole model. To facilitate discussions in Section \ref{subsec:fixed}, we draw a definition of the interpretation degree in the model.
\begin{definition}
Define $\phi(x \vert M)$ as the interpretation degree of observation $x$ in the G-components mixture model, which also represents the maximum density of $x$ in a single model component. $\phi(x \vert M)$ can be given by:
\begin{equation}\label{eq:phi_x_m}
  \phi(x \vert M)=max \lbrace \phi_1(x \vert \theta_1), \phi_2(x \vert \theta_2), \cdots, \phi_G(x \vert \theta_G)\rbrace.
\end{equation}
\end{definition}

\subsection{A Fixed Point View}\label{subsec:fixed}
Model-based clustering algorithm mainly relies on the continuous reuse of the data and the iterative updating of the model parameters until convergence. A classical iterative scheme for model-based clustering is the EM algorithm. By iterating, the data observation $X$ can match one or more components in the mixed model, which corresponds to the iterative changes of $\phi(x \vert M)$ values. Typically, as the number of iterations increases, $\phi(x \vert M)$ of data observation tends to become $1$. For example, in the Gaussian mixture model algorithm, the $\mu$ parameter in the model indicates the center of the model component. A more general condition has been taken into consideration in this article that, the parameters do not indicate the the clustering centers. Instead, it is the fixed points that suggest the clustering center, which are formalized by the contraction map. In order to ensure the compressing characteristic of the contraction map, the following two definitions are proposed first.

\begin{definition}
Define $\alpha$-sequence as an incremental sequence with time serial $t$ satisfying: $\exists C\in (0,1]$ and $C$ is constant, so that:
\begin{equation}
\left\{
\begin{aligned}
&\alpha^{(t)} < \alpha^{(t+1)} <C, \forall t \geq 0&\\
&\alpha^{(0)} =0&\\
&\lim_{t \to \infty} \alpha^{(t)} = C&
\end{aligned}
\right.
.
\end{equation}
\end{definition}

\begin{definition}
Define $\alpha$-critical space as a subspace $S$ in $R^L$. S is the convex hull of data observation satisfying $\phi(x \vert M)\geq \alpha$. This can be interpreted as:
\begin{equation}
S(\alpha)=Conv \lbrace x \vert \phi(x \vert M)\geq \alpha \rbrace.
\end{equation}
In particular, if $S$ is an $\alpha$-critical space in space $D$, then we have:
\begin{equation*}
S_D (\alpha)=Conv \lbrace x \vert \phi(x \vert M)\geq \alpha,x \in D \rbrace.
\end{equation*}
\end{definition}

When the model-based clustering algorithm iterates to the $t^{th}$ time, there always exists $t$ and $\alpha^{(t)}$ so that $S_U(\alpha^{(t)}) \neq \emptyset$. To simplify the notation, $S_U(\alpha^{(t)})$ is denoted as $S^{(t)}$. The current nonempty $S^{(t)}$ can be regarded as the universal set for constructing $\alpha$-critical space, if and only if there exist $x_0$ and $\epsilon >0$ so that $Borel(x_0,\epsilon)\subset S^{(t)}$.
From the above analysis, it is worthy of note that a spatial mapping can always be constructed, due to the adjustment of parameters before and after the iteration. We clarify the compression of the space as a mapping for better explaining and analyzing the mechanism.

\begin{definition}
For a given model component $g$, before the iteration, the nonempty universal set is $S^{(i-1)}$, which is generated after the $(i-1)^{th}$ iteration. For the $i^{th}$ iteration:
\begin{enumerate}[1)]
\item If there exist $x_0$ and $\epsilon >0$ so that $Borel(x_0,\epsilon)\subset S^{(i-1)}$, then find an adequate $\alpha^{(i)}$ so that $S^{(i)} \subset S^{(i-1)}$. Define a surjection $H_g^i$ for the $g^{th}$ model component at the $i^{th}$ iteration as $H_g^i:S^{(i-1)}\to S^{(i)}$. In this article, the surjection is called the contraction map, which also called the $H$-map in the following discussions.
\item If the condition in 1) is not satisfied, the model component converges.
\end{enumerate}
\end{definition}

\begin{figure}[htbp]
\centering
\includegraphics[width=0.6\textwidth]{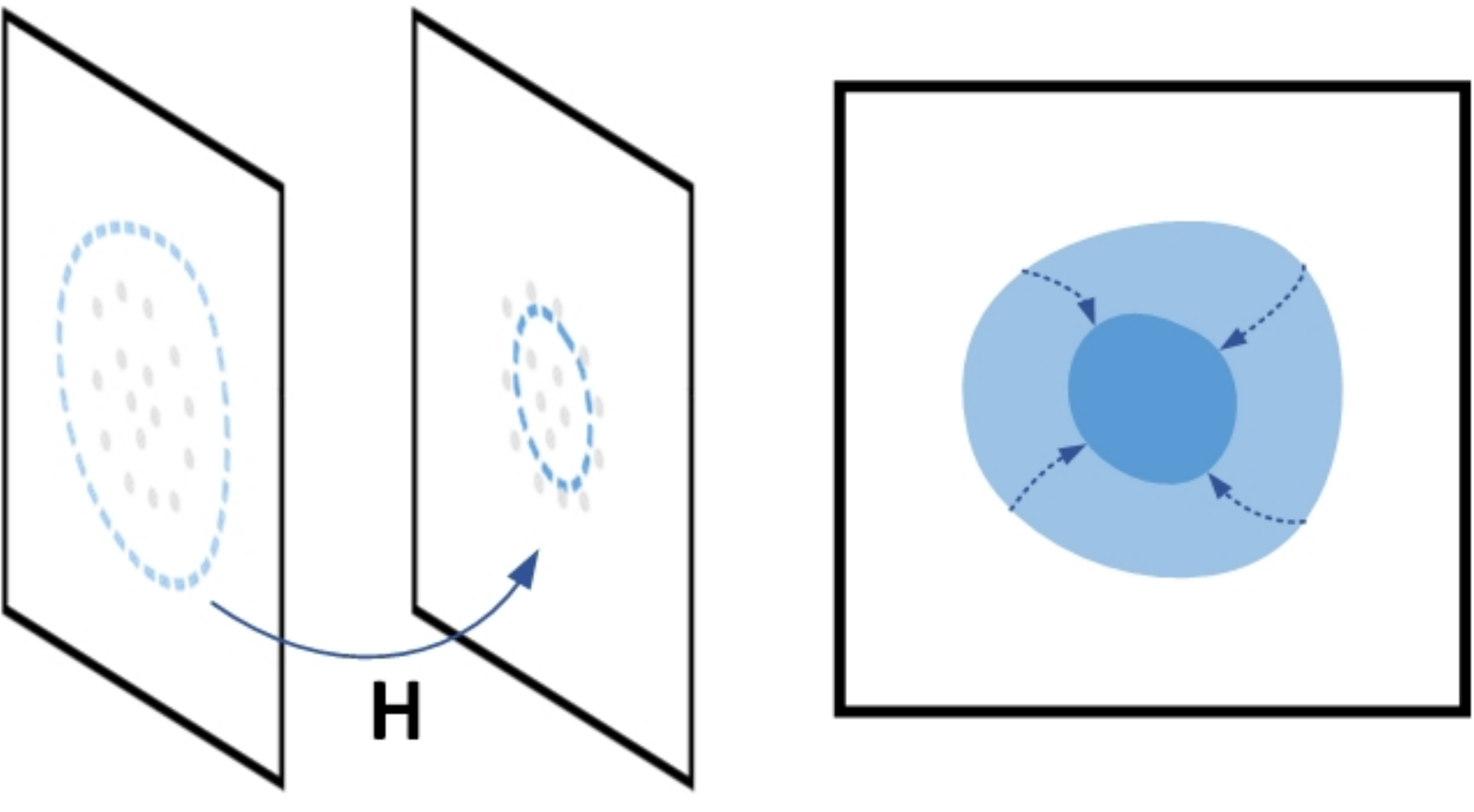}
\caption{The construction of the H-map from the iteration of the algorithm}
\label{fig:h_map}
\end{figure}

Note that for the $i^{th}$ iteration, by finding an adequate $\alpha^{(i)}$, we construct a compressed space $S^{(i)}$ in $S^{(i-1)}$. The contraction of $S^{(i)}$ guarantee the existence of fix points for the $H$-map, which ensures the converge of algorithm in finite time. When $S^{(i)}$ is compressed to a single observation point, that is, diameter $diam(S^{(i)})=0$, the $H$-map discovers the only fixed point. And when all the $H$-maps in the algorithm find the unique fixed point, the algorithm complete. Thus, the clusters of the model can be discriminated by the number of fixed points and the structure of the model.

In actual fact, almost all the convergence and discriminant process of clustering algorithms can be explained by the fixed points of the $H$-map. Although they represent different meanings in different algorithms, fixed points can be generally interpreted as the cluster centers. In common clustering algorithms such as K-means and GMM, fixed points are usually the center of their data distribution. It is because of the utilization of central iterative schemes, as the parameters of the model priori are also centric. But for other algorithms, such as DBSCAN \cite{MAHESHKUMAR201639,Ester421283}, OPTICS \cite{Ankerst1999}, DENCLUE \cite{Hinneburg1998}, they are variants from density-based clustering. These algorithms are identical in idea, which is identifying a high-density region as a cluster and a density center as a cluster center. Owing to the difference between density-based feature and distance-based feature, their fixed points are rarely consistent with the distance-based center.

Notice that the $H$-map does not guarantee the uniqueness of the fixed point. In other words, more than one fixed point may exist in the compressed space, which constitutes the structure of tree. In the proposed unified framework, the iteration objectively ensures the existence of $\alpha^{(t)}$ less than the constant $C$ to achieve the compression before convergence. The growth of the $\alpha$-sequence can assess the suitability of the model. Note that each contraction map can only be mapped to a simply-connected $\alpha$-critical space. The segmentation of space should be conducted for a more complex space such as multiply-connected space or a closed manifold.

The most important idea of the proposed framework is that, during the iteration of the clustering, the cluster center is given by abstracting the map and determining the feasible subspace, rather than directly characterized by the parameters. With the iteration of the model, the feasible subspace for fixed points shrinks with the growth of $\alpha$-sequence. In other words, the convergence of clustering is a necessary condition of satisfying the framework in this paper.

By this point, the basic form of the unified clustering framework has been introduced. The methodology of framework is shown in Algorithm 1.

\begin{algorithm}[tb]
    \caption{A Unified Framework for model-based clustering}
    \begin{algorithmic}[1]
        \REQUIRE Data observation $X$, Mixture models $M$ ($G_0$ components)
        \REQUIRE Initial parameter $\theta_0$, Initial mixture coefficient $\pi_0$
        \REQUIRE Step size $\Delta\alpha$
        \ENSURE First term of $\alpha$-sequence $\alpha^{(0)}=0$, Convex hull $S^{(0)}=Conv(X)$
        \ENSURE $G^{(0)}=G_0$, $\theta^{(0)}=\theta_0$, $\pi^{(0)}=\pi_0$, $i=0$
        \WHILE{not converged}
        \STATE Adjust $\theta^{(i)}$, $\pi^{(i)}$ using estimation of observation
        \STATE Adjust $\phi(x \vert M)\leftarrow max\lbrace \phi_1 (x \vert \theta_1 ),\phi_2 (x \vert \theta_2 ),\cdots,\phi_G (x \vert \theta_G)\rbrace $
        \STATE Adjust $\alpha^{(i+1)}\leftarrow \Delta\alpha + \alpha^{(i)}$
        \STATE Adjust all $S_g^{(i+1)} \leftarrow H(S_g^i )  (g=1,2,\cdots,G^{(0)})$
        \STATE Adjust $G^{(i+1)}$ in accordance with the number of $S_g^{(i+1)}$
        \IF {$diam(S_g^{(i+1)})=diam(S_g^{(i)})$}
        \STATE The algorithm converges
        \ELSE
        \STATE $i \leftarrow i+1$
        \ENDIF
        \ENDWHILE
        \PRINT Mixture models $M$
    \end{algorithmic}
\end{algorithm}

\section{Relation with GMM}\label{sec:extended}

GMM is a typical Model-Based Clustering Algorithm. Its parameter update relies on the EM estimation or the Maximum A Posteriori (MAP) method. Suppose that there are $G_0$ predicted components, then components will converge independently to $G_0$ fixed points of the corresponding space. Therefore, GMM is mapped to the framework with the EM estimation method to discuss the application of framework.

Given an $L$-dimensional observation set $X$, and assume that the data distribution is subjected to the Gaussian distribution and each component is independent. The estimation steps get the similarity between the Gaussian distribution and the data by the maximum likelihood method. The posterior probabilities for the components is calculated based on the initial or previous parameters. Denote $\phi_g (X_i \vert \theta_g )$ as the posterior probability.

The posterior probability of $X_i$ in the $g^{th}$ model component is given by \cite{Reynolds2009}:
\begin{equation}
\phi_g ( X_i \vert \theta_g^{(t)}) =\frac{\pi_g^{(t)} \mathcal{N}(X_i \vert \mu_g^{(t)},\Sigma_g^{(t)} )}{\sum_{k=1}^G \pi_k^{(t)} \mathcal{N}(X_i \vert \mu_k^{(t)},\Sigma_k^{(t)} )}.
\end{equation}

Then, the mixture coefficient $\pi^{(t+1)}$, the covariance matrix $\Sigma^{(t+1)}$, and the model centers $\mu^{(t+1)}$ in step $t+1$ are updated based on the posterior probability in the previous step:
\begin{equation*}
\begin{split}
\pi_g^{(t+1)}&=\frac{\sum_{i=1}^N \phi_g(X_i \vert \theta_g^{(t)})}{N};\\
\Sigma^{(t+1)}&=\frac{\sum_{i=1}^N \phi_g(X_i \vert \theta_g^{(t)}) (X_i-\mu_g^{(t)}) (X_i-\mu_g^{(t)})^T}{\sum_{i=1}^N \phi_g(X_i \vert \theta_g^{(t)})};\\
\mu^{(t+1)}&=\frac{\sum_{i=1}^N \phi_g(X_i \vert \theta_g^{(t)}) X_i}{\sum_{i=1}^N \phi_g(X_i \vert \theta_g^{(t)})}
\end{split}
\end{equation*}

Select one dimension (denoted as $e$) of the observation vector for further study. Due to the features of the Gaussian distribution, the distribution in the observational dimension reduce from the multi-dimensional Gaussian distribution to one-dimensional Gaussian distribution $f_e(x \vert \pi,\theta)$ ($\sigma_{e,g}^{(t)}$ and $\mu_{e,g}^{(t)}$ are parameters in observational dimension for the $g^{th}$ component):
\begin{equation}
f_e(x \vert \pi,\theta)=\sum_{g=1}^{G_0} \frac{\pi_g}{\sqrt{2\pi} \sigma_{e,g}^{(t)}} exp(- \frac{{\vert x-\mu_{e,g}^{(t)} \vert}^2}{2{\sigma_{e,g}^{(t)}}^2})
\end{equation}

The convex hull $Conv(X)$ is also projected into closed interval $Proj[Conv(X)]$ in observational dimension, which satisfies $Proj[Conv(X)] \subset \Gamma_e$. $\Gamma_e$ is the $\alpha$-critical space in observational dimension. For the $g^{th}$ model component:
\begin{equation*}
\Gamma_{e,g}^{(t)}=\lbrace x \vert \mu_{e,g}^{(t)} - \sigma_{e,g}^{(t)} \sqrt{-2 ln(\sqrt{2\pi} \alpha^{(t)} \sigma_{e,g}^{(t)})} \leq x \leq \mu_{e,g}^{(t)} + \sigma_{e,g}^{(t)} \sqrt{-2 ln(\sqrt{2\pi} \alpha^{(t)} \sigma_{e,g}^{(t)})} \rbrace
\end{equation*}

To ensure the result of $\sqrt{-2 ln(\sqrt{2\pi} \alpha^{(t)} \sigma_{e,g}^{(t)})}$ is real number, $\alpha^{(t)}<1 / \sqrt{2\pi} \alpha^{(t)} \sigma_{e,g}^{(t)}$ should be satisfied. Let the length of interval be $l_{e,g}=2 \sigma_{e,g}^{(t)} \sqrt{-2 ln(\sqrt{2\pi} \alpha^{(t)} \sigma_{e,g}^{(t)})}$ and obtain the partial derivative of $l_g$ with regard to $\alpha$:
\begin{equation*}
\frac{\partial l_{e,g}}{\partial \alpha}=-\frac{{4\sigma_{e,g}^{(t)}}^2}{\alpha l_{e,g}}.
\end{equation*}

As $l_{e,g}>0$ and $\sigma_{e,g}^{(t)}>0$, $\partial l_{e,g}/\partial \alpha$ is negative constantly, so the interval $\Gamma_{e,g}^{(t)}$ shrinks as $\alpha$ increases.

Considering $\Gamma_{e,g}^{(t)}$ and $\Gamma_{e,g}^{(t+1)}$, if $\Gamma_{e,g}^{(t)} \cap \Gamma_{e,g}^{(t+1)}\neq \emptyset$, let K be $\frac{sup(\Gamma_{e,g}^{(t)} \cap \Gamma_{e,g}^{(t+1)})-inf(\Gamma_{e,g}^{(t)} \cap \Gamma_{e,g}^{(t+1)})}{sup(\Gamma_{e,g}^{(t)})-inf(\Gamma_{e,g}^{(t)})}$, from the above analyses, it is apparent that $K\in [0,1)$. The contracting map $H_g$ in observational dimension $H_{e,g}: \Gamma_{e,g}^{(t)} \to \Gamma_{e,g}^{(t+1)}$ can be constructed as:
\begin{equation}
y=K(x-inf(\Gamma_{e,g}^{(t)})).
\end{equation}

Then for all $x_1,x_2\in \Gamma_{e,g}^{(t)}$, $\Vert H_{e,g}(x_1)-H_{e,g}(x_2)\Vert \leq K\Vert x_1-x_2\Vert$, $K \in[0,1)$, which satisfies the Lipschitz condition. By the Banach fixed point theorem, $H_{e,g}$ map is proved to have only one fixed point.

As a result, the framework guarantees the convergence under certain conditions. Under the domination of the number of clusters, the nonempty intersection of the previous and the latter $\alpha$-critical space may be achieved and a linear contraction map can be constructed between the previous and the latter $\alpha$-critical space. If the above conditions are met, the framework can ensure that, for each component, no extra fixed point will be generated. This, hence, maintain the priori not to be changed by iteration. The reasonable explanation for the above conclusion is that, the iterative method has absolute or logarithmic linear characteristic, which makes it reasonable to construct the linear mapping to satisfy the Lipschitiz condition. Similar analyses can also be applied to the distance-based clustering method for further discussion.

\begin{figure}[htbp]
  \centering
  \subfloat[]
  {\includegraphics[width=0.3\textwidth,trim={130 40 130 40},clip]{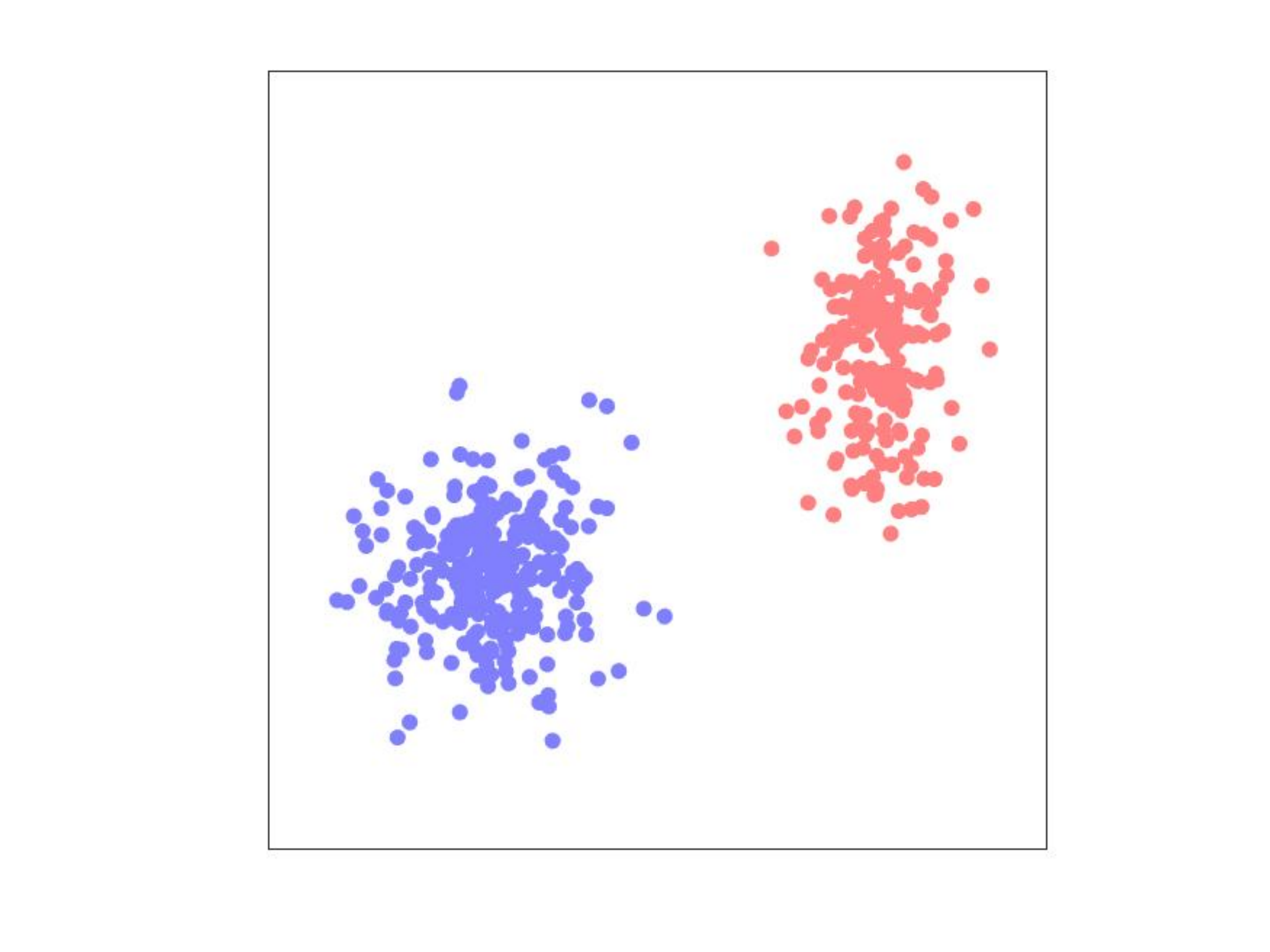}}
  \quad
  \subfloat[]
  {\includegraphics[width=0.3\textwidth,trim={130 40 130 40},clip]{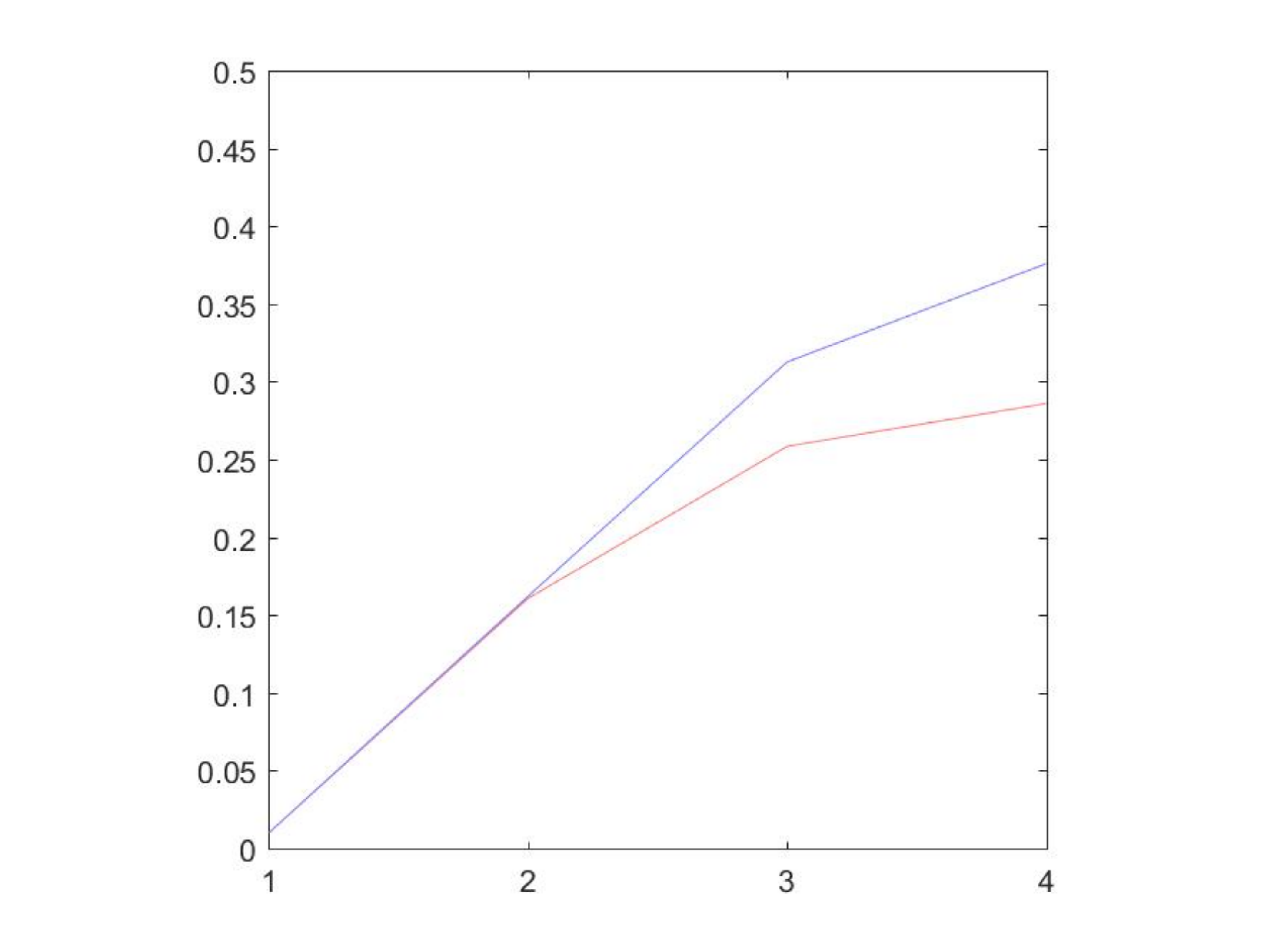}}
  \quad
  \subfloat[]
  {\includegraphics[width=0.3\textwidth,trim={130 40 130 40},clip]{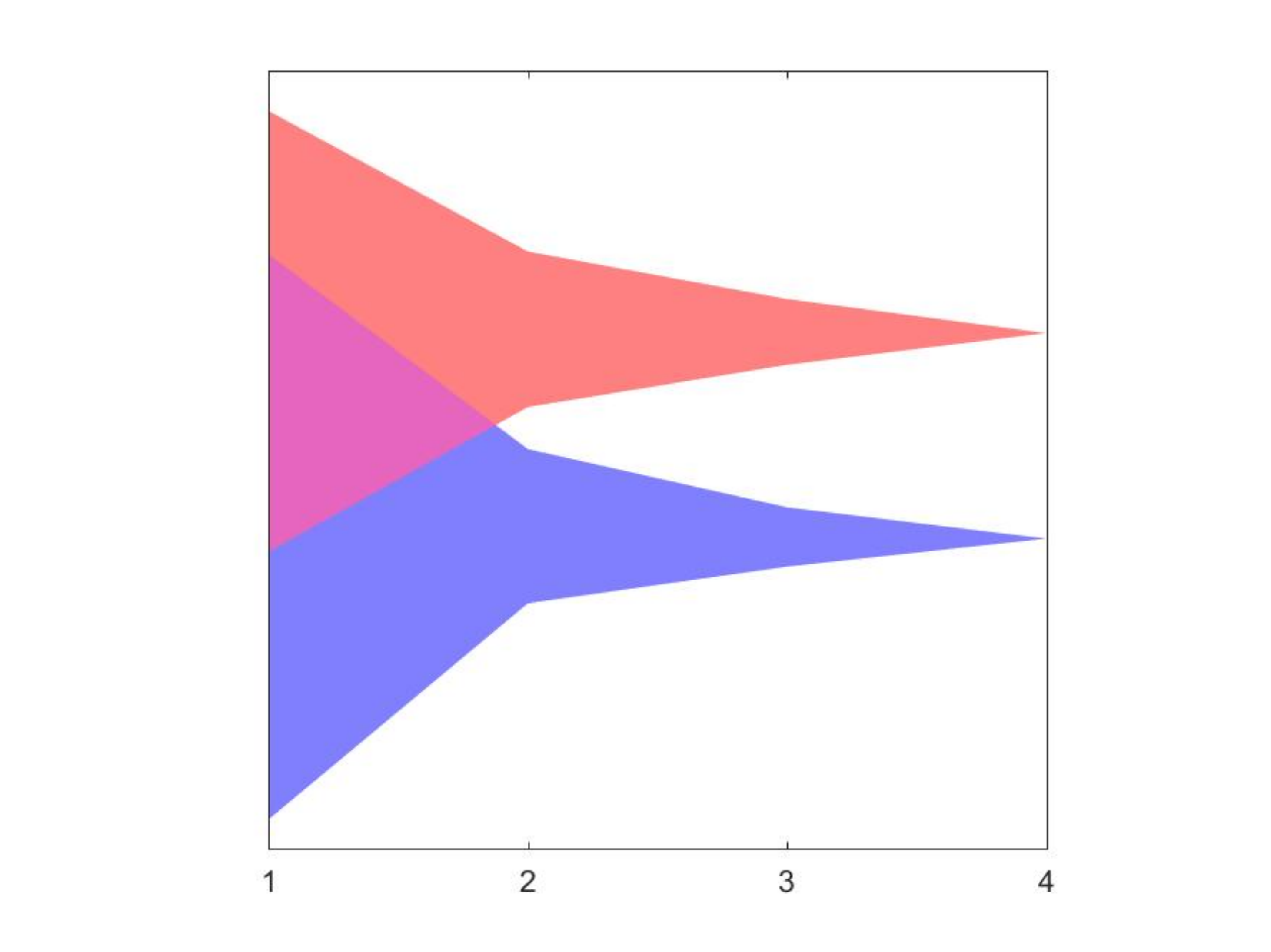}}
  \caption{The scatter diagram of a Gaussian mixture model with two components (a) and the visualization of the $\alpha$-sequence (b) and the contraction map (c) of two components in one dimension during the iteration.}
  \label{fig:testify}
\end{figure}

An experiment is carried out. By embedding the GMM algorithm, we adjust the $\alpha$-sequence, for the purpose of compressing the $\alpha$-critical space of the different components. The upper bound of the $\alpha$-sequence is related to the variance of the component. According to the above analysis, a smaller variance is related to a greater upper bound. Figure \ref{fig:testify} presents the visualization of the experiment.

\section{Conclusion and Future Work}\label{sec:conclusion}
In this paper, after the analysis on the model-based clustering, a unified framework for model-based clustering is proposed from the view of fixed point. It iteratively constructs the contraction map to find fixed points as cluster centers. Through specifying a contraction map, the GMM algorithm is mapped to the proposed framework as an application. The framework can inspire us to develop future modular clustering algorithm. It can be applied in representative learning and achieve unsupervised recognition and reasoning. 


\newpage
\section*{Reference}
\bibliographystyle{elsarticle-num}
\bibliography{main}

\begin{thebibliography}{10}
\expandafter\ifx\csname url\endcsname\relax
  \def\url#1{\texttt{#1}}\fi
\expandafter\ifx\csname urlprefix\endcsname\relax\def\urlprefix{URL }\fi
\expandafter\ifx\csname href\endcsname\relax
  \def\href#1#2{#2} \def\path#1{#1}\fi

\bibitem{Sowmya2017}
S.~R, S.~K. R, Data mining with big data, in: 2017 11th International
  Conference on Intelligent Systems and Control (ISCO), 2017, pp. 246--250.
\newblock \href {http://dx.doi.org/10.1109/ISCO.2017.7855990}
  {\path{doi:10.1109/ISCO.2017.7855990}}.

\bibitem{Xu2015}
D.~Xu, Y.~Tian, \href{https://doi.org/10.1007/s40745-015-0040-1}{A
  comprehensive survey of clustering algorithms}, Annals of Data Science 2~(2)
  (2015) 165--193.
\newblock \href {http://dx.doi.org/10.1007/s40745-015-0040-1}
  {\path{doi:10.1007/s40745-015-0040-1}}.
\newline\urlprefix\url{https://doi.org/10.1007/s40745-015-0040-1}

\bibitem{NERURKAR2018770}
P.~Nerurkar, A.~Shirke, M.~Chandane, S.~Bhirud,
  \href{http://www.sciencedirect.com/science/article/pii/S1877050917328673}{Empirical
  analysis of data clustering algorithms}, Procedia Computer Science 125 (2018)
  770 -- 779, the 6th International Conference on Smart Computing and
  Communications.
\newblock \href {http://dx.doi.org/https://doi.org/10.1016/j.procs.2017.12.099}
  {\path{doi:https://doi.org/10.1016/j.procs.2017.12.099}}.
\newline\urlprefix\url{http://www.sciencedirect.com/science/article/pii/S1877050917328673}

\bibitem{mcnicholas2016model}
P.~D. McNicholas, Model-based clustering, Journal of Classification 33~(3)
  (2016) 331--373.

\bibitem{Jin2017}
X.~Jin, J.~Han,
  \href{https://doi.org/10.1007/978-1-4899-7687-1_344}{Expectation Maximization
  Clustering}, Springer US, Boston, MA, 2017, pp. 480 -- 482.
\newblock \href {http://dx.doi.org/10.1007/978-1-4899-7687-1_344}
  {\path{doi:10.1007/978-1-4899-7687-1_344}}.
\newline\urlprefix\url{https://doi.org/10.1007/978-1-4899-7687-1_344}

\bibitem{Reynolds2009}
D.~Reynolds, \href{https://doi.org/10.1007/978-0-387-73003-5_196}{Gaussian
  Mixture Models}, Springer US, Boston, MA, 2009, pp. 659--663.
\newblock \href {http://dx.doi.org/10.1007/978-0-387-73003-5_196}
  {\path{doi:10.1007/978-0-387-73003-5_196}}.
\newline\urlprefix\url{https://doi.org/10.1007/978-0-387-73003-5_196}

\bibitem{CARPENTER1990129}
G.~A. Carpenter, S.~Grossberg,
  \href{http://www.sciencedirect.com/science/article/pii/089360809090085Y}{Art
  3: Hierarchical search using chemical transmitters in self-organizing pattern
  recognition architectures}, Neural Networks 3~(2) (1990) 129 -- 152.
\newblock \href
  {http://dx.doi.org/https://doi.org/10.1016/0893-6080(90)90085-Y}
  {\path{doi:https://doi.org/10.1016/0893-6080(90)90085-Y}}.
\newline\urlprefix\url{http://www.sciencedirect.com/science/article/pii/089360809090085Y}

\bibitem{Carpenter198712}
G.~A.~Carpenter, S.~Grossberg, Art 2: Stable self-organization of pattern
  recognition codes for analog input patterns 26 (1987) 4919--30.
\newblock \href {http://dx.doi.org/10.1364/AO.26.004919}
  {\path{doi:10.1364/AO.26.004919}}.

\bibitem{CARPENTER198754}
G.~A. Carpenter, S.~Grossberg,
  \href{http://www.sciencedirect.com/science/article/pii/S0734189X87800142}{A
  massively parallel architecture for a self-organizing neural pattern
  recognition machine}, Computer Vision, Graphics, and Image Processing 37~(1)
  (1987) 54 -- 115.
\newblock \href {http://dx.doi.org/10.1016/S0734-189X(87)80014-2}
  {\path{doi:10.1016/S0734-189X(87)80014-2}}.
\newline\urlprefix\url{http://www.sciencedirect.com/science/article/pii/S0734189X87800142}

\bibitem{Zhong2003}
S.~Zhong, J.~Ghosh, \href{http://dl.acm.org/citation.cfm?id=945365.964287}{A
  unified framework for model-based clustering}, J. Mach. Learn. Res. 4 (2003)
  1001--1037.
\newblock \href {http://dx.doi.org/10.1162/1532443041827943}
  {\path{doi:10.1162/1532443041827943}}.
\newline\urlprefix\url{http://dl.acm.org/citation.cfm?id=945365.964287}

\bibitem{Teboulle2007}
M.~Teboulle, \href{http://dl.acm.org/citation.cfm?id=1248659.1248662}{A unified
  continuous optimization framework for center-based clustering methods},
  Journal of Machine Learning Research 8 (2007) 65--102.
\newline\urlprefix\url{http://dl.acm.org/citation.cfm?id=1248659.1248662}

\bibitem{Scellier2017}
B.~Scellier, Y.~Bengio,
  \href{http://europepmc.org/articles/PMC5415673}{Equilibrium propagation:
  Bridging the gap between energy-based models and backpropagation}, Frontiers
  in computational neuroscience 11 (2017) 24.
\newblock \href {http://dx.doi.org/10.3389/fncom.2017.00024}
  {\path{doi:10.3389/fncom.2017.00024}}.
\newline\urlprefix\url{http://europepmc.org/articles/PMC5415673}

\bibitem{Kleinberg2003}
J.~Kleinberg, An impossibility theorem for clustering 15 (2003) 446--453.
\newblock \href {http://dx.doi.org/10.1103/PhysRevE.90.062813}
  {\path{doi:10.1103/PhysRevE.90.062813}}.

\bibitem{Zadeh2009}
R.~B. Zadeh, S.~Ben-David,
  \href{http://dl.acm.org/citation.cfm?id=1795114.1795189}{A uniqueness theorem
  for clustering}, in: Proceedings of the Twenty-Fifth Conference on
  Uncertainty in Artificial Intelligence, UAI '09, AUAI Press, Arlington,
  Virginia, United States, 2009, pp. 639--646.
\newline\urlprefix\url{http://dl.acm.org/citation.cfm?id=1795114.1795189}

\bibitem{Bandyopadhyay2016}
S.~Bandyopadhyay, M.~N. Murty, Axioms to characterize efficient incremental
  clustering, in: 2016 23rd International Conference on Pattern Recognition
  (ICPR), 2016, pp. 450--455.
\newblock \href {http://dx.doi.org/10.1109/ICPR.2016.7899675}
  {\path{doi:10.1109/ICPR.2016.7899675}}.

\bibitem{KlopotekK2017}
R.~A. Klopotek, M.~A. Klopotek, \href{http://arxiv.org/abs/1702.04577}{On the
  discrepancy between kleinberg's clustering axioms and k-means clustering
  algorithm behavior}, CoRR abs/1702.04577.
\newblock \href {http://arxiv.org/abs/1702.04577} {\path{arXiv:1702.04577}}.
\newline\urlprefix\url{http://arxiv.org/abs/1702.04577}

\bibitem{Li2015}
C.-G. Li, R.~Vidal, Structured sparse subspace clustering: A unified
  optimization framework, in: 2015 IEEE Conference on Computer Vision and
  Pattern Recognition (CVPR), 2015, pp. 277--286.
\newblock \href {http://dx.doi.org/10.1109/CVPR.2015.7298624}
  {\path{doi:10.1109/CVPR.2015.7298624}}.

\bibitem{Yang2016}
Y.~Yang, F.~Shen, Z.~Huang, H.~T. Shen,
  \href{http://dl.acm.org/citation.cfm?id=3060832.3060939}{A unified framework
  for discrete spectral clustering}, in: Proceedings of the Twenty-Fifth
  International Joint Conference on Artificial Intelligence, IJCAI'16, AAAI
  Press, 2016, pp. 2273--2279.
\newline\urlprefix\url{http://dl.acm.org/citation.cfm?id=3060832.3060939}

\bibitem{McNicholas2016}
P.~D. McNicholas, \href{https://doi.org/10.1007/s00357-016-9211-9}{Model-based
  clustering}, Journal of Classification 33~(3) (2016) 331--373.
\newblock \href {http://dx.doi.org/10.1007/s00357-016-9211-9}
  {\path{doi:10.1007/s00357-016-9211-9}}.
\newline\urlprefix\url{https://doi.org/10.1007/s00357-016-9211-9}

\bibitem{MAHESHKUMAR201639}
K.~M. Kumar, A.~R.~M. Reddy,
  \href{http://www.sciencedirect.com/science/article/pii/S0031320316001035}{A
  fast dbscan clustering algorithm by accelerating neighbor searching using
  groups method}, Pattern Recognition 58 (2016) 39 -- 48.
\newblock \href {http://dx.doi.org/10.1016/j.patcog.2016.03.008}
  {\path{doi:10.1016/j.patcog.2016.03.008}}.
\newline\urlprefix\url{http://www.sciencedirect.com/science/article/pii/S0031320316001035}

\bibitem{Ester421283}
M.~Ester, H.~Kriegel, J.~Sander, X.~Xiaowei, A density-based algorithm for
  discovering clusters in large spatial databases with noise (1996) 226--231.

\bibitem{Ankerst1999}
M.~Ankerst, M.~M. Breunig, H.-P. Kriegel, J.~Sander,
  \href{http://doi.acm.org/10.1145/304181.304187}{Optics: Ordering points to
  identify the clustering structure}, SIGMOD Rec. 28~(2) (1999) 49--60.
\newblock \href {http://dx.doi.org/10.1145/304181.304187}
  {\path{doi:10.1145/304181.304187}}.
\newline\urlprefix\url{http://doi.acm.org/10.1145/304181.304187}

\bibitem{Hinneburg1998}
A.~Hinneburg, D.~A. Keim,
  \href{http://dl.acm.org/citation.cfm?id=3000292.3000302}{An efficient
  approach to clustering in large multimedia databases with noise}, in:
  Proceedings of the Fourth International Conference on Knowledge Discovery and
  Data Mining, KDD'98, AAAI Press, 1998, pp. 58--65.
\newline\urlprefix\url{http://dl.acm.org/citation.cfm?id=3000292.3000302}

\end{thebibliography}

\end{document}